\pdfoutput=1

\documentclass[11pt]{article}

\usepackage[]{EMNLP2022}

\usepackage{times}
\usepackage{latexsym}

\usepackage{multirow}
\usepackage{subcaption}
\usepackage{graphicx}
\usepackage{diagbox}
\usepackage{adjustbox}
\usepackage{float}
\usepackage{stfloats}
\usepackage{hyperref}

\usepackage[T1]{fontenc}

\usepackage[utf8]{inputenc}

\usepackage{microtype}

\usepackage{inconsolata}

\usepackage{rotating}

\title{An Exploration of Data Efficiency\\ in Intra-Dataset Task Transfer for Dialog Understanding}

\author{Josiah Ross, Luke Yoffe, Alon Albalak, William Yang Wang\\
  University of California, Santa Barbara \\ 
  }

\begin{document}
\maketitle
\begin{abstract}
Transfer learning is an exciting area of Natural Language Processing that has the potential to both improve model performance and increase data efficiency. This study explores the effects of varying quantities of target task training data on sequential transfer learning in the dialog domain. We hypothesize that a model can utilize the information learned from a source task to better learn a target task, thereby reducing the number of target task training samples required. Unintuitively, our data shows that often target task training data size has minimal effect on how sequential transfer learning performs compared to the same model without transfer learning\footnote{We used \href{https://github.com/josiahnross/TLiDB}{github.com/josiahnross/TLiDB} to run our experiments}. Our results lead us to believe that this unexpected result could be due to the effects of catastrophic forgetting, motivating further work into methods that prevent such forgetting.

\end{abstract}

\section{Introduction}
Large annotated datasets are needed to train state-of-the-art NLP models. Models pretrained on self-supervised language modeling tasks such as T5 \cite{JMLR:v21:20-074} and BERT \cite{devlin-etal-2019-bert} have improved accuracy and data efficiency on downstream tasks,
and it has been demonstrated that this
concept can be pushed further through supervised task transfer \cite{pruksachatkun-etal-2020-intermediate}.

Transfer learning is a technique where a machine learning model can use the knowledge it has learned from one task or domain in order to better perform in a different task or domain \cite{Pan2010ASO}. This study explores task transfer (transfer learning between tasks), specifically intra-dataset task transfer, where the source and target tasks are annotated on the same dataset \cite{https://doi.org/10.48550/arxiv.2205.06262}. We decided to focus on intra-dataset task transfer in order to specifically study the effect of varying the amount of target task data used on task transfer without allowing our results to get affected by switching the domain. 


We hope that our work and future work on task transfer can make NLP more accessible and efficient. Task transfer allows larger institutions to use their vast resources to train models such as BERT on a supervised task and publish the resulting model for others to start from.  For example, if someone wanted to train a model to perform reading comprehension, it might be better to start from a BERT model already trained on emotion recognition than the base BERT model.

We hypothesize that a model needs less target task data in order to have similar performance to a model trained without task transfer since, through task transfer, a model can use what it learns from the source task to learn the target task more efficiently. Additionally, we theorize that when a model has access to large quantities of target task data, transfer learning would be less effective, since the target task data would contain enough knowledge on its own. Our goal for this paper is to explore intra-dataset task transfer's effect on data efficiency and determine whether or not our theories are correct. Contrary to our hypothesis, our results show that sequential intra-domain task transfer doesn't necessarily improve data efficiency.

\section{Dataset, Models, and Framework}
In this section, we describe the dataset, models, and framework of our experiments. Many of our decisions regarding our choice of dataset, models, and our focus on intra-dataset task transfer are based on the FETA paper \cite{https://doi.org/10.48550/arxiv.2205.06262}. Additionally, we based our code off of \citet{Albalak_The_Transfer_Learning_2022}. However, unlike \citet{https://doi.org/10.48550/arxiv.2205.06262}, who ran their experiments using only 10\% of the target task training and validation data, we ran our experiments on 20\%, 40\%, 60\%, 80\% and 100\% of the target task training and validation data.  This allows us to observe the effect of transfer learning on target task data efficiency.
\subsection{Transfer Method}
Task transfer can be accomplished in various ways, but we decided to use sequential task transfer \cite{MCCLOSKEY1989109, ruder-etal-2019-transfer}. The sequential method involves training the model in two distinct steps: (1) training the model on a source task, then (2) training the model on the target task. What differentiates the sequential method from other methods, such as multitask and multitask/fine-tuning transfer learning \cite{10.5555/2998687.2998769}, is that the model never trains on the source and target task at the same time.  This is ideal because it would allow large institutions that are less limited by computing power and resources to annotate data, train large models, and then publish those models. Then, smaller groups can use these published source task models as the base for their target task models. Since the training of the model on the source and target tasks is in separate steps, this is sequential learning.

\subsection{Dataset}
We use the \textit{Friends} dialog dataset (see Appendix \ref{appendix:friends} for details) for our experiments. 70\% of the dialogues in the dataset are used for training, 15\% for validation, and 15\% for testing.
From there the training and validation dialogues are divided into 20\%, 40\%, 60\%, 80\%, and 100\% data splits such that every dialog in a smaller split appears in each larger split. For example, a dialog in the 40\% training split must also appear in the 60\%, 80\% and 100\% splits but may or may not appear in the 20\% split. In this way, larger data divisions are guaranteed to have at least as much information as smaller data divisions, even if certain dialogues inherently hold more information. Each sample isn't annotated on each task so some tasks have more samples than others. Additionally, the data split samples are the same across all tasks meaning that if a dialog is in the testing data split it's also in the testing data split for every other task it is annotated on. See Appendix \ref{appendix:DataSplitDetails} for exact data split counts.

\subsection{Models}
To see if results differed across pretrained models, we ran our experiments on both BERT-base from \citet{devlin-etal-2019-bert} (110 million parameters) and T5-base from \citet{JMLR:v21:20-074} (220 million parameters). On top of the base BERT model a small classification layer, unique to each task, was trained to convert the output of the base model to a valid output of the task. T5 converts all tasks into a text-to-text format and requires no additional classification layer.

\section{Experiment Methodology}

\subsection{Evaluation and Stopping Mechanism}
Each task has one or more metrics it can be evaluated on (see Appendix \ref{appendix:FriendsTasks} for metric details). When a model is evaluated on a task with multiple metrics, its performance is based on that model's average performance over that task's metrics.

For each experiment, training stopped when the validation metric hadn't improved in 4 consecutive epochs. The validation metric was calculated using the same percentage of the validation data as the was used for the training data. For instance, when training on the 40\% training split the model will be validation metric will be calculated on the 40\% validation split. We saved the model checkpoint that had the highest validation metric. Then, after the training was complete, we evaluated the best model checkpoint on 100\% of the testing data (regardless of the percentage of the data used during training) and used that score in our analysis.

\subsection{Hyperparameter Search}
For each task on each pretrained model type, we ran a hyperparameter search across the learning rates $10^{-4}$ and $10^{-5}$ and the batch sizes 10, 30, 60, and 120.  We used 100\% of the data for training and validation and did not use task transfer. Every two epochs, the hyperparameter combinations were evaluated using 100\% of the validation data and the worst 45\% were removed so that the search didn't waste time training hyperparameters already known to be subpar. The search stopped after every hyperparameter combination had either been removed or stopped by the stopping mechanism. Once finished, the hyperparameter search saved the hyperparameter combination that had the highest validation metric on 100\% of the validation data. These saved hyperparameters were then used in our experiments regardless of whether or not task transfer was used or which percentage of the data was being used for training. See Appendix \ref{appendix:Hyperparameters} for the results of the hyperparameter search.

\subsection{Source and Target Task Models}
Once the hyperparameters were determined, the source models were trained on each task using both BERT and T5. The source models were all trained using 100\% of the training data. We decided not to alter the number of samples each source task model used to train because we wanted to focus on the data efficiency of the target tasks rather than the data efficiency of the source tasks. Since we were exploring sequential transfer learning, the source model only needs to be trained once but could be used many times on different target tasks. Therefore, the data efficiency on the source task is much less important than the data efficiency on the target task.

Additionally, we chose not to equalize the number of samples used in each task. Some tasks are cheaper to annotate than others (per sample), and we made the assumption that the dataset was created by spending an equal amount of resources to annotate each task.  Under this assumption, any differences in the number of samples between two tasks can be attributed to the difference in cost per sample of annotating those tasks.
This assumption only matters when we are comparing different source tasks, it doesn't affect the trends we see within a given source task.

After creating models trained on each source task, we then trained a copy of each of these source task models on every other task at every percentage (20/40/60/80/100) of the target task data. These models are our target task models. 

\subsection{No Transfer Models}
As well as training the source models on 100\% of the training data, we also trained models on each task with both BERT and T5 without task transfer using 20, 40, 60, and 80\% of the data. We ran this experiment so that we could measure the effect of task transfer by comparing the performance of a model trained using task transfer to the performance of a model trained using the same data but with no task transfer.




\subsection{Masked Language Modeling}
Since the pretrained models we are using (BERT and T5) were pretrained on a more general dataset than what we are using, we wanted to look into the effect of continuing unsupervised learning on our dataset, which has previously shown positive results in specific domains \cite{10.1145/3458754}.  BERT was pretrained in part using masked language modeling, where 15\% of the input tokens are masked and the model is tasked with predicting what the masked tokens were.  

To test if unsupervised learning on our dataset would help, we added masked language modeling to our list of BERT source tasks. Masked language modeling is unsupervised, so we were able to simply use every dialog in each data split as our training, validation, and testing data (without having to worry about annotating the data). Notably, the 15\% of the input tokens that were masked were re-chosen randomly at the beginning of each epoch.

\subsection{Random Seeds}
Both the initialization of the models and the training of the models relied on some randomness. We ran each of our experiments 3 times using a different training/initialization seed each time and we averaged all results across the 3 seeds. Crucially, the data splits were generated randomly but remained the same across all experiments.

\begin{figure*}[t]
    \centering
    \includegraphics[width=\textwidth]{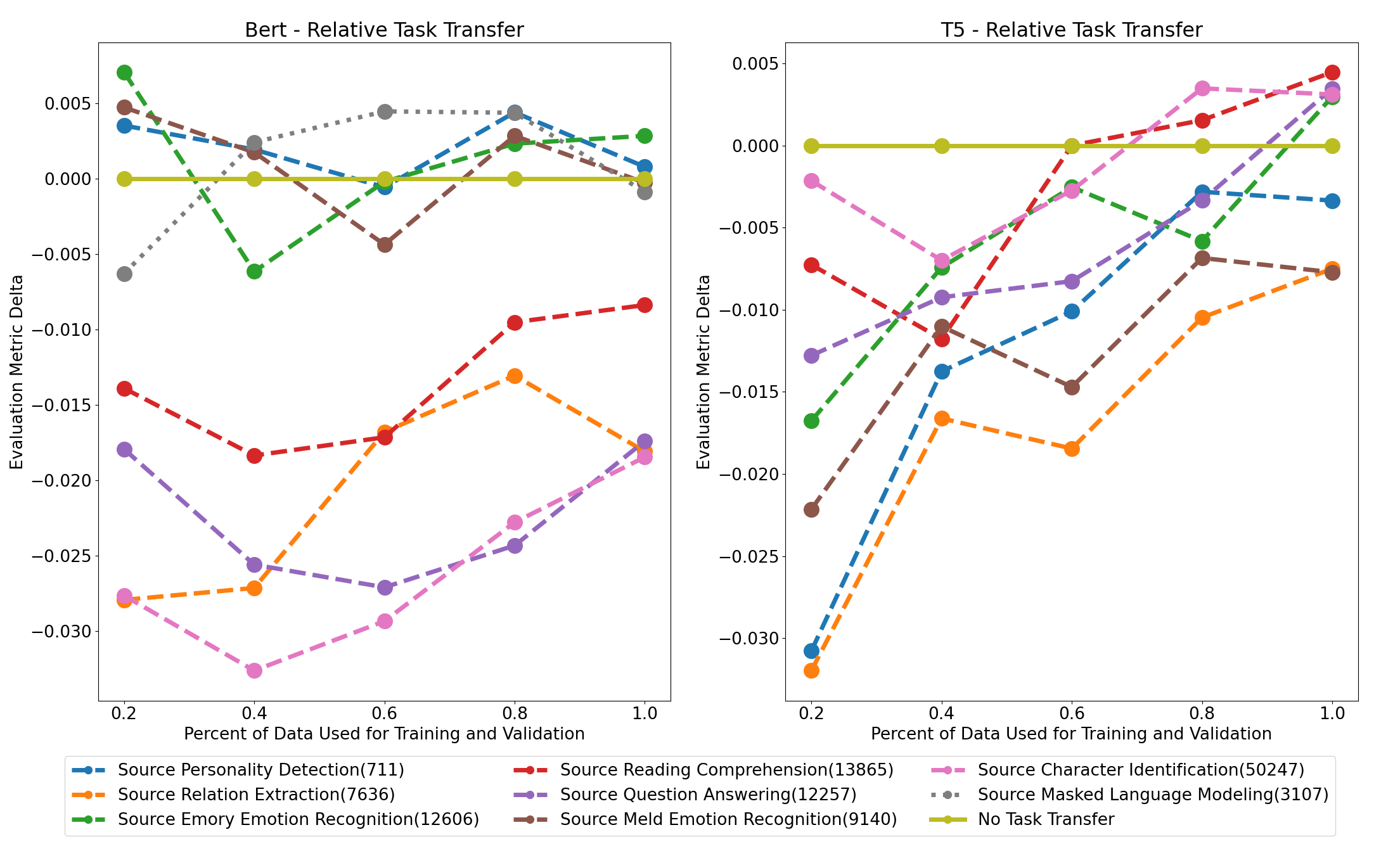}
    \caption{How each source task performed on average on the BERT (Left) and T5 (Right) models. The x-axis represents which percentage of the data was used for training and validation. The y-axis represents the average difference between the target task performance using task transfer and target task performance without task transfer. The performance of each source task is averaged over each valid target task (every other task besides masked language modeling) and over the 3 training/initialization random seeds.}
    \label{fig:MainResults}
\end{figure*}

\section{Results}
We originally hypothesized that we would see a downward trend in our results showing that when less target task data is used, transfer learning is more effective. However, this is not what Figure \ref{fig:MainResults} shows. Our BERT results show a fairly flat trend while our T5 results show an upward trend. We hypothesize that these trends are due to catastrophic forgetting \cite{MCCLOSKEY1989109, DBLP:journals/corr/abs-1801-01423}. When the model learns the source task, it forgets some of what it learned during its pretraining. To compensate for this forgetting, the model needs a large amount of target task training data to relearn any of the forgotten knowledge important to modeling the target task. This isn't a problem if the model isn't performing task transfer since the model will tend to only forget things that aren't relevant to the current task. If this theory is true, we would see the difference in scores between transfer and no transfer increase as the percent of the target task training data increases.

While that is what we see in our T5 results, we don't see this in our BERT results. We hypothesize that there are two forces at play in task transfer:
\begin{enumerate}
    \item When training on a source task, some new knowledge is learned, leading to better results in low-data settings.
    \item When training on a source task, some knowledge learned during pretraining is forgotten, leading to worse results in low-data settings.
\end{enumerate}
In the case of T5, (2) may be having a stronger effect leading to the general upward trend, while in the case of BERT, these effects are more equal, so they cancel each other out resulting in a mostly flat trend.


Our results show that, for many source tasks, task transfer produces a negative effect. While unfortunate, this result is somewhat expected since sequential task transfer is particularly susceptible to the effects of catastrophic forgetting \cite{MCCLOSKEY1989109}. In the case of BERT, there seem to be two distinct clusters of source tasks in terms of their performance, but T5 does not have this pattern.  This suggests that different model architectures are affected by learning and forgetting in task transfer differently.


Interestingly, while masked language modeling appears to be one of the best BERT source tasks, it still had very little effect in comparison to not using task transfer. This contradicts our expectation that masked language modeling would always help by adapting the BERT model from the domain it was pretrained on to the \textit{Friends} domain. We hypothesize that the reason why we don't see this result is simply because BERT's pretraining is general enough to understand the dialogues, so performing masked language modeling on our domain has little effect.

\section{Conclusion}
In this paper, we explored intra-dataset sequential task transfer’s effect on target efficiency. We found that with BERT, adjusting the amount of target task data used did not significantly affect task transfer's effectiveness. With T5, we found that the more target task data we used, the more effective task transfer was. Additionally, we found that out of our 7 tasks, Personality Detection, Emory Emotion Recognition, MELD Emotion Recognition, and Masked Language Modeling on BERT had a neutral effect in comparison to not using transfer learning, while the remaining tasks on BERT and T5 had mostly negative results. We also showed that just because a source task performs better than other tasks on BERT that doesn't mean it will on T5 or vice versa.
We suspect that our counter-intuitive results are due to the effects of catastrophic forgetting and we encourage future research into catastrophic forgetting and its effects on transfer learning.


\section{Impact Statement}
The experiments in this study have used a considerable amount of computing power, so it is important that we consider the energy and environmental costs of this work. In total, we trained about 1750 models, each taking about 55 GPU minutes on average to train and evaluate on NVIDIA TITAN X GPUs. This took a total of about 1586 GPU hours or 66 GPU days total. While this is certainly a large amount of computing time, the goal of this paper is to reduce the amount of computing time needed in the long run. While our results might not have not directly led to this outcome, we hope that future studies can use our paper as a stepping stone to improve data efficiency, reducing both energy and annotation costs. Successful task transfer can reduce the training time needed as one source task model can be reused as a starting point for many different target task models.

\section{Limitations}
While we hoped to make our study as comprehensive as possible, there were some limitations to our experiments. First, we ran each experiment with only three different model initialization and training seeds, which means our results could be affected by variations between the seeds. Second, we only produced our data splits (20/40/60/80/100\% of data) once and reused those splits in all of our experiments. Ideally, we would have re-run our experiments using several different data splits. Next, we used one dataset in English for our experiments that included 7 different annotated tasks. Ideally, we would have used several datasets of differing languages each with many annotated tasks. Next, in this study, we decided to focus on intra-dataset transfer. However, we hope that similar experiments across domains will be done in the future, as domain adaptation will be important when starting with a source model trained on a different dataset. Another limitation of this study is that we only used the sequential transfer learning method. We would like to see other methods of transfer learning such as multitask and multitask/fine-tune \cite{10.5555/2998687.2998769} explored in future work. Lastly, this study didn't explore the effect of altering the amount of source task samples used, but this is another variable that could be altered to produce useful results in future work.

\bibliography{anthology,custom}

\begin{thebibliography}{19}
\expandafter\ifx\csname natexlab\endcsname\relax\def\natexlab#1{#1}\fi

\bibitem[{Albalak(2022)}]{Albalak_The_Transfer_Learning_2022}
Alon Albalak. 2022.
\newblock \href {https://doi.org/10.5281/zenodo.6374360} {{The Transfer
  Learning in Dialogue Benchmarking Toolkit}}.

\bibitem[{Albalak et~al.(2022)Albalak, Tuan, Jandaghi, Pryor, Yoffe,
  Ramachandran, Getoor, Pujara, and
  Wang}]{https://doi.org/10.48550/arxiv.2205.06262}
Alon Albalak, Yi-Lin Tuan, Pegah Jandaghi, Connor Pryor, Luke Yoffe, Deepak
  Ramachandran, Lise Getoor, Jay Pujara, and William~Yang Wang. 2022.
\newblock \href {https://doi.org/10.48550/ARXIV.2205.06262} {Feta: A benchmark
  for few-sample task transfer in open-domain dialogue}.

\bibitem[{Caruana(1994)}]{10.5555/2998687.2998769}
Rich Caruana. 1994.
\newblock Learning many related tasks at the same time with backpropagation.
\newblock In \emph{Proceedings of the 7th International Conference on Neural
  Information Processing Systems}, NIPS'94, page 657–664, Cambridge, MA, USA.
  MIT Press.

\bibitem[{Chen and Choi(2016)}]{chen-choi-2016-character}
Yu-Hsin Chen and Jinho~D. Choi. 2016.
\newblock \href {https://doi.org/10.18653/v1/W16-3612} {Character
  identification on multiparty conversation: Identifying mentions of characters
  in {TV} shows}.
\newblock In \emph{Proceedings of the 17th Annual Meeting of the Special
  Interest Group on Discourse and Dialogue}, pages 90--100, Los Angeles.
  Association for Computational Linguistics.

\bibitem[{Devlin et~al.(2019)Devlin, Chang, Lee, and
  Toutanova}]{devlin-etal-2019-bert}
Jacob Devlin, Ming-Wei Chang, Kenton Lee, and Kristina Toutanova. 2019.
\newblock \href {https://doi.org/10.18653/v1/N19-1423} {{BERT}: Pre-training of
  deep bidirectional transformers for language understanding}.
\newblock In \emph{Proceedings of the 2019 Conference of the North {A}merican
  Chapter of the Association for Computational Linguistics: Human Language
  Technologies, Volume 1 (Long and Short Papers)}, pages 4171--4186,
  Minneapolis, Minnesota. Association for Computational Linguistics.

\bibitem[{Gu et~al.(2021)Gu, Tinn, Cheng, Lucas, Usuyama, Liu, Naumann, Gao,
  and Poon}]{10.1145/3458754}
Yu~Gu, Robert Tinn, Hao Cheng, Michael Lucas, Naoto Usuyama, Xiaodong Liu,
  Tristan Naumann, Jianfeng Gao, and Hoifung Poon. 2021.
\newblock \href {https://doi.org/10.1145/3458754} {Domain-specific language
  model pretraining for biomedical natural language processing}.
\newblock \emph{ACM Trans. Comput. Healthcare}, 3(1).

\bibitem[{Jiang et~al.(2019)Jiang, Zhang, and
  Choi}]{DBLP:journals/corr/abs-1911-09304}
Hang Jiang, Xianzhe Zhang, and Jinho~D. Choi. 2019.
\newblock \href {http://arxiv.org/abs/1911.09304} {Automatic text-based
  personality recognition on monologues and multiparty dialogues using
  attentive networks and contextual embeddings}.
\newblock \emph{CoRR}, abs/1911.09304.

\bibitem[{Ma et~al.(2018)Ma, Jurczyk, and Choi}]{ma-etal-2018-challenging}
Kaixin Ma, Tomasz Jurczyk, and Jinho~D. Choi. 2018.
\newblock \href {https://doi.org/10.18653/v1/N18-1185} {Challenging reading
  comprehension on daily conversation: Passage completion on multiparty
  dialog}.
\newblock In \emph{Proceedings of the 2018 Conference of the North {A}merican
  Chapter of the Association for Computational Linguistics: Human Language
  Technologies, Volume 1 (Long Papers)}, pages 2039--2048, New Orleans,
  Louisiana. Association for Computational Linguistics.

\bibitem[{McCloskey and Cohen(1989)}]{MCCLOSKEY1989109}
Michael McCloskey and Neal~J. Cohen. 1989.
\newblock \href {https://doi.org/https://doi.org/10.1016/S0079-7421(08)60536-8}
  {Catastrophic interference in connectionist networks: The sequential learning
  problem}.
\newblock volume~24 of \emph{Psychology of Learning and Motivation}, pages
  109--165. Academic Press.

\bibitem[{Pan and Yang(2010)}]{Pan2010ASO}
Sinno~Jialin Pan and Qiang Yang. 2010.
\newblock A survey on transfer learning.
\newblock \emph{IEEE Transactions on Knowledge and Data Engineering},
  22:1345--1359.

\bibitem[{Poria et~al.(2018)Poria, Hazarika, Majumder, Naik, Cambria, and
  Mihalcea}]{DBLP:journals/corr/abs-1810-02508}
Soujanya Poria, Devamanyu Hazarika, Navonil Majumder, Gautam Naik, Erik
  Cambria, and Rada Mihalcea. 2018.
\newblock \href {http://arxiv.org/abs/1810.02508} {{MELD:} {A} multimodal
  multi-party dataset for emotion recognition in conversations}.
\newblock \emph{CoRR}, abs/1810.02508.

\bibitem[{Pruksachatkun et~al.(2020)Pruksachatkun, Phang, Liu, Htut, Zhang,
  Pang, Vania, Kann, and Bowman}]{pruksachatkun-etal-2020-intermediate}
Yada Pruksachatkun, Jason Phang, Haokun Liu, Phu~Mon Htut, Xiaoyi Zhang,
  Richard~Yuanzhe Pang, Clara Vania, Katharina Kann, and Samuel~R. Bowman.
  2020.
\newblock \href {https://doi.org/10.18653/v1/2020.acl-main.467}
  {Intermediate-task transfer learning with pretrained language models: When
  and why does it work?}
\newblock In \emph{Proceedings of the 58th Annual Meeting of the Association
  for Computational Linguistics}, pages 5231--5247, Online. Association for
  Computational Linguistics.

\bibitem[{Raffel et~al.(2020)Raffel, Shazeer, Roberts, Lee, Narang, Matena,
  Zhou, Li, and Liu}]{JMLR:v21:20-074}
Colin Raffel, Noam Shazeer, Adam Roberts, Katherine Lee, Sharan Narang, Michael
  Matena, Yanqi Zhou, Wei Li, and Peter~J. Liu. 2020.
\newblock \href {http://jmlr.org/papers/v21/20-074.html} {Exploring the limits
  of transfer learning with a unified text-to-text transformer}.
\newblock \emph{Journal of Machine Learning Research}, 21(140):1--67.

\bibitem[{Ruder et~al.(2019)Ruder, Peters, Swayamdipta, and
  Wolf}]{ruder-etal-2019-transfer}
Sebastian Ruder, Matthew~E. Peters, Swabha Swayamdipta, and Thomas Wolf. 2019.
\newblock \href {https://doi.org/10.18653/v1/N19-5004} {Transfer learning in
  natural language processing}.
\newblock In \emph{Proceedings of the 2019 Conference of the North {A}merican
  Chapter of the Association for Computational Linguistics: Tutorials}, pages
  15--18, Minneapolis, Minnesota. Association for Computational Linguistics.

\bibitem[{Serr{\`{a}} et~al.(2018)Serr{\`{a}}, Sur{\'{\i}}s, Miron, and
  Karatzoglou}]{DBLP:journals/corr/abs-1801-01423}
Joan Serr{\`{a}}, D{\'{\i}}dac Sur{\'{\i}}s, Marius Miron, and Alexandros
  Karatzoglou. 2018.
\newblock \href {http://arxiv.org/abs/1801.01423} {Overcoming catastrophic
  forgetting with hard attention to the task}.
\newblock \emph{CoRR}, abs/1801.01423.

\bibitem[{Yang and Choi(2019)}]{yang-choi-2019-friendsqa}
Zhengzhe Yang and Jinho~D. Choi. 2019.
\newblock \href {https://doi.org/10.18653/v1/W19-5923} {{F}riends{QA}:
  Open-domain question answering on {TV} show transcripts}.
\newblock In \emph{Proceedings of the 20th Annual SIGdial Meeting on Discourse
  and Dialogue}, pages 188--197, Stockholm, Sweden. Association for
  Computational Linguistics.

\bibitem[{Yu et~al.(2020)Yu, Sun, Cardie, and
  Yu}]{DBLP:journals/corr/abs-2004-08056}
Dian Yu, Kai Sun, Claire Cardie, and Dong Yu. 2020.
\newblock \href {http://arxiv.org/abs/2004.08056} {Dialogue-based relation
  extraction}.
\newblock \emph{CoRR}, abs/2004.08056.

\bibitem[{Zahiri and Choi(2017)}]{DBLP:journals/corr/abs-1708-04299}
Sayyed~M. Zahiri and Jinho~D. Choi. 2017.
\newblock \href {http://arxiv.org/abs/1708.04299} {Emotion detection on {TV}
  show transcripts with sequence-based convolutional neural networks}.
\newblock \emph{CoRR}, abs/1708.04299.

\bibitem[{Zhou and Choi(2018)}]{zhou-choi-2018-exist}
Ethan Zhou and Jinho~D. Choi. 2018.
\newblock \href {https://aclanthology.org/C18-1003} {They exist! introducing
  plural mentions to coreference resolution and entity linking}.
\newblock In \emph{Proceedings of the 27th International Conference on
  Computational Linguistics}, pages 24--34, Santa Fe, New Mexico, USA.
  Association for Computational Linguistics.

\end{thebibliography}
\bibliographystyle{acl_natbib}

\appendix

\section{\textit{Friends} Dataset Details}
\label{appendix:friends}
\subsection{Tasks}
\label{appendix:FriendsTasks}
The \textit{Friends} dataset contains dialogues from the TV show \textit{Friends} annotated on the following 7 tasks.

\textbf{Character Identification} is the task of mapping mentions of characters in a multi-party dialog to the correct characters \cite{chen-choi-2016-character}. Evaluated on the Macro-F1 and Micro-F1 metrics. Annotations provided by \citet{chen-choi-2016-character} and \citet{zhou-choi-2018-exist}.

\textbf{Emory Emotion Recognition} is the task of assigning one of 7 emotions to each utterance in a dialogue. Evaluated on the Micro-F1 and Weighted-F1 metrics. Annotations provided by \citet{DBLP:journals/corr/abs-1708-04299}.

\textbf{MELD Emotion Recognition} is the task of assigning one of 7 emotions to each utterance in a dialogue. While clearly similar to \textbf{Emory Emotion Recognition}, \textbf{MELD Emotion Recognition} uses 2 different emotions and only 22\% of the dialogues annotated on \textbf{Emory Emotion Recognition} and \textbf{MELD Emotion Recognition} overlap. Evaluated on the Micro-F1 and Weighted-F1 metrics. Annotations provided by \citet{DBLP:journals/corr/abs-1810-02508}.

\textbf{Reading Comprehension} is the task of answering fill in the blank factual questions about a dialogue. Evaluated on the Accuracy metric. Annotations provided by \citet{ma-etal-2018-challenging} 

\textbf{Question Answering} is the task of answering a question about a dialogue. The answer to each question is always either a span within the dialog or a speaker. Evaluated on the Token-F1 and Exact Match Metrics. Annotations provided by \citet{yang-choi-2019-friendsqa}.

\textbf{Personality Detection} is the task of determining whether a speaker in a dialogue exhibits any of 5 different personality traits. Evaluated on the Accuracy metric. Annotations provided by \citet{DBLP:journals/corr/abs-1911-09304}

\textbf{Relation Extraction} is the task of predicting the relationship between two subjects in a dialogue. Evaluated on the Micro-F1 metric. Annotations provided by \citet{DBLP:journals/corr/abs-2004-08056}. 

\begin{figure*}[b!]
    \includegraphics[width=\textwidth]{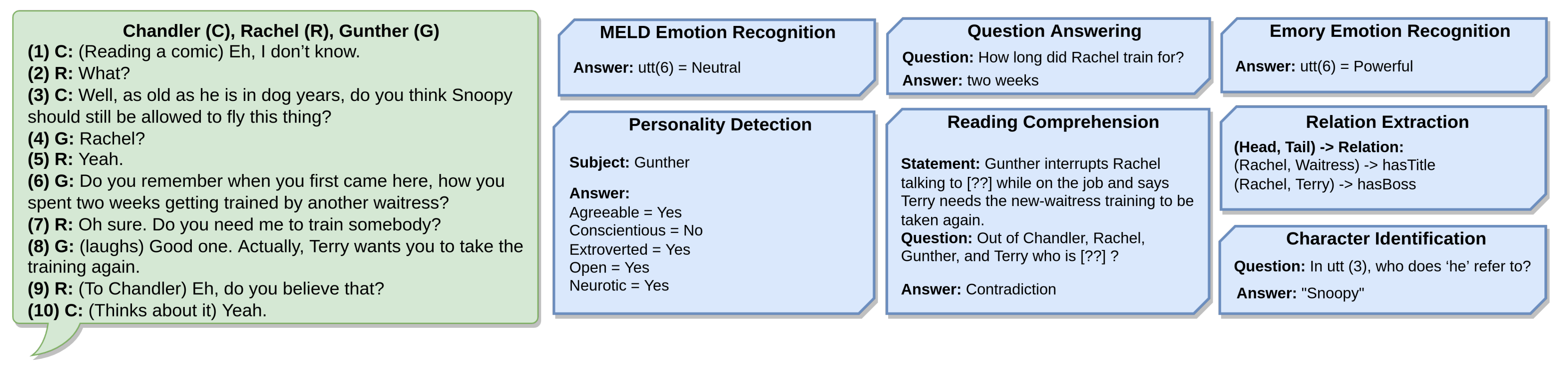}
    \caption{Example \textit{Friends} dialogue with annotated tasks. Source: \citet{https://doi.org/10.48550/arxiv.2205.06262}.}
    \label{fig:my_label3}
\end{figure*}
\clearpage

\label{appendix:DataSplitDetails}
\begin{sidewaystable}[b!]
\centering
\begin{tabular}{|c|||c||c||c|c|c|c|c||c|c|c|c|c|}
\hline
\multicolumn{3}{|c|}{} & 20\% & 40\% & 60\% & 80\% & 100\%  & 20\% & 40\% & 60\% & 80\% & 100\% \\ \hline 
Task & Total & Testing & \multicolumn{5}{|c||}{Training} & \multicolumn{5}{|c|}{Validation} \\ \hline \hline
Personality Detection & 711 & 110 & 103 & 212 & 317 & 387 & 487 & 29 & 53 & 75 & 96 & 114\\ \hline 
Relation Extraction & 7312 & 1131 & 1098 & 2068 & 3070 & 4065 & 4953 & 211 & 458 & 739 & 1026 & 1228\\ \hline 
Emory Emotion Recognition & 12606 & 1912 & 1712 & 3572 & 5180 & 6926 & 8629 & 474 & 853 & 1263 & 1651 & 2065\\ \hline 
Reading Comprehension & 13865 & 2284 & 2095 & 3945 & 5845 & 7735 & 9503 & 455 & 936 & 1298 & 1790 & 2078\\ \hline 
Question Answering & 12257 & 1937 & 1633 & 3368 & 4983 & 6592 & 8351 & 399 & 763 & 1218 & 1569 & 1969\\ \hline 
Character Identification & 50247 & 7803 & 7519 & 14603 & 21353 & 27848 & 34724 & 1727 & 3367 & 4925 & 6323 & 7720\\ \hline 
Meld Emotion Recognition & 9140 & 1247 & 1353 & 2690 & 4072 & 5390 & 6603 & 344 & 553 & 809 & 1062 & 1290\\ \hline 
Masked Language Modeling & 3107 & 466 & 435 & 870 & 1305 & 1740 & 2175 & 93 & 186 & 280 & 373 & 466\\ \hline
\end{tabular}
\caption{Amount of samples in each data split for each target task.}
\end{sidewaystable}
\subsection{Dataset Split Details}
\clearpage

\section{Hyperparameters Search Results}
\label{appendix:Hyperparameters}
\begin{table*}[hb]
\centering
\begin{tabular}{|c||c|c|c|}
\hline
Task & Learning Rate & Batch Size \\ \hline \hline
Personality Detection & $10^{-5}$ & 60\\ \hline 
Relation Extraction & $10^{-5}$ & 10\\ \hline 
Emory Emotion Recognition & $10^{-5}$ & 30\\ \hline 
Reading Comprehension & $10^{-5}$ & 10\\ \hline 
Question Answering & $10^{-4}$ & 120\\ \hline 
Character Identification & $10^{-5}$ & 10\\ \hline 
Meld Emotion Recognition & $10^{-5}$ & 30\\ \hline 
Masked Language Modeling & $10^{-5}$ & 10\\ \hline
\end{tabular}
\caption{Learning rate and batch sizes used when training tasks on BERT.}
\end{table*}
\begin{table*}[hb]
\centering
\begin{tabular}{|c||c|c|c|}
\hline
Task & Learning Rate & Batch Size \\ \hline \hline
Personality Detection & $10^{-4}$ & 120\\ \hline 
Relation Extraction & $10^{-4}$ & 10\\ \hline 
Emory Emotion Recognition & $10^{-4}$ & 120\\ \hline 
Reading Comprehension & $10^{-4}$ & 60\\ \hline 
Question Answering & $10^{-5}$ & 10\\ \hline 
Character Identification & $10^{-5}$ & 10\\ \hline 
Meld Emotion Recognition & $10^{-4}$ & 10\\ \hline 
\end{tabular}
\caption{Learning rate and batch sizes used when training tasks on T5.}
\end{table*}

\clearpage

\section{Expanded Results}
\label{appendix:results}

\begin{table*}[hb]
\centering
\begin{tabular}{|l|c|c|c|c|c|c|}
\hline
Percentage of Target Task Data & 20\% & 40\% & 60\% & 80\% & 100\%\\ \hline
\multicolumn{6}{|c|}{Target Task: Personality Detection on BERT}\\ \hline
No Task Transfer Personality Detection & 0.4848 & 0.5406 & 0.537 & 0.5467 & 0.5539\\
 & (0.0438) & (0.0056) & (0.0126) & (0.0073) & (0.023)\\ \hline
Source Task: Relation Extraction & 0.5273 & 0.5539 & 0.5442 & 0.5485 & 0.5509\\
 & (0.0065) & (0.0034) & (0.006) & (0.0155) & (0.0079)\\ \hline
Source Task: Emory Emotion Recognition & 0.5273 & 0.5315 & 0.543 & 0.5455 & 0.5491\\
 & (0.0054) & (0.0095) & (0.0087) & (0.0104) & (0.0182)\\ \hline
Source Task: Reading Comprehension & 0.5491 & 0.5539 & 0.5679 & 0.5642 & 0.5648\\
 & (0.0225) & (0.0082) & (0.0082) & (0.0209) & (0.012)\\ \hline
Source Task: Question Answering & 0.4976 & 0.5582 & 0.5327 & 0.5448 & 0.5412\\
 & (0.0256) & (0.0157) & (0.0195) & (0.0115) & (0.0105)\\ \hline
Source Task: Character Identification & 0.5358 & 0.5479 & 0.5515 & 0.5527 & 0.5503\\
 & (0.0037) & (0.0177) & (0.0149) & (0.0026) & (0.0031)\\ \hline
Source Task: Meld Emotion Recognition & 0.517 & 0.537 & 0.5364 & 0.5497 & 0.5461\\
 & (0.0184) & (0.0141) & (0.0129) & (0.0056) & (0.0109)\\ \hline
Source Task: Masked Language Modeling & 0.4848 & 0.5406 & 0.537 & 0.5467 & 0.5539\\
 & (0.0438) & (0.0056) & (0.0126) & (0.0073) & (0.023)\\ \hline
\end{tabular}
\caption{Mean and standard deviation (in parentheses) of the performance of each source task transferred to the target task Personality Detection for each percentage of the target task data on BERT.}
\end{table*}

\begin{table*}
\centering
\begin{tabular}{|l|c|c|c|c|c|c|}
\hline
Percentage of Target Task Data & 20\% & 40\% & 60\% & 80\% & 100\%\\ \hline
\multicolumn{6}{|c|}{Target Task: Relation Extraction on BERT}\\ \hline
No Task Transfer Relation Extraction & 0.3603 & 0.4951 & 0.5545 & 0.573 & 0.5794\\
 & (0.0449) & (0.0116) & (0.0219) & (0.0105) & (0.0139)\\ \hline
Source Task: Personality Detection & 0.3957 & 0.511 & 0.5558 & 0.5628 & 0.6022\\
 & (0.01) & (0.0126) & (0.0153) & (0.0113) & (0.0269)\\ \hline
Source Task: Emory Emotion Recognition & 0.35 & 0.462 & 0.5356 & 0.5519 & 0.5826\\
 & (0.021) & (0.0186) & (0.0169) & (0.0096) & (0.0145)\\ \hline
Source Task: Reading Comprehension & 0.3349 & 0.4063 & 0.4837 & 0.5336 & 0.5524\\
 & (0.0275) & (0.0127) & (0.0105) & (0.0172) & (0.0013)\\ \hline
Source Task: Question Answering & 0.355 & 0.4017 & 0.4699 & 0.4844 & 0.5576\\
 & (0.02) & (0.0297) & (0.0212) & (0.0142) & (0.0341)\\ \hline
Source Task: Character Identification & 0.3612 & 0.4696 & 0.4993 & 0.5256 & 0.5769\\
 & (0.0286) & (0.0419) & (0.0591) & (0.0293) & (0.0322)\\ \hline
Source Task: Meld Emotion Recognition & 0.365 & 0.4887 & 0.5305 & 0.5524 & 0.5957\\
 & (0.0132) & (0.0187) & (0.0294) & (0.0243) & (0.0055)\\ \hline
Source Task: Masked Language Modeling & 0.3603 & 0.4951 & 0.5545 & 0.573 & 0.5794\\
 & (0.0449) & (0.0116) & (0.0219) & (0.0105) & (0.0139)\\ \hline
\end{tabular}
\caption{Mean and standard deviation (in parentheses) of the performance of each source task transferred to the target task Relation Extraction for each percentage of the target task data on BERT.}
\end{table*}

\begin{table*}
\centering
\begin{tabular}{|l|c|c|c|c|c|c|}
\hline
Percentage of Target Task Data & 20\% & 40\% & 60\% & 80\% & 100\%\\ \hline
\multicolumn{6}{|c|}{Target Task: Emory Emotion Recognition on BERT}\\ \hline
No Task Transfer Emory Emotion Recognition & 0.3072 & 0.3549 & 0.3723 & 0.3859 & 0.3935\\
 & (0.0347) & (0.0097) & (0.0031) & (0.0154) & (0.006)\\ \hline
Source Task: Personality Detection & 0.3385 & 0.374 & 0.3804 & 0.3857 & 0.4014\\
 & (0.011) & (0.0052) & (0.012) & (0.008) & (0.0103)\\ \hline
Source Task: Relation Extraction & 0.2637 & 0.3121 & 0.3375 & 0.3485 & 0.3543\\
 & (0.0143) & (0.0201) & (0.0028) & (0.0057) & (0.0092)\\ \hline
Source Task: Reading Comprehension & 0.3093 & 0.3491 & 0.3554 & 0.3803 & 0.3863\\
 & (0.0183) & (0.0178) & (0.0179) & (0.0038) & (0.0034)\\ \hline
Source Task: Question Answering & 0.3057 & 0.3496 & 0.3553 & 0.3594 & 0.3823\\
 & (0.0099) & (0.01) & (0.0105) & (0.0107) & (0.0108)\\ \hline
Source Task: Character Identification & 0.2397 & 0.2456 & 0.29 & 0.3022 & 0.3322\\
 & (0.0131) & (0.0323) & (0.0371) & (0.0443) & (0.0176)\\ \hline
Source Task: Meld Emotion Recognition & 0.3556 & 0.3771 & 0.3876 & 0.3929 & 0.4012\\
 & (0.0076) & (0.0161) & (0.0046) & (0.002) & (0.0073)\\ \hline
Source Task: Masked Language Modeling & 0.3072 & 0.3549 & 0.3723 & 0.3859 & 0.3935\\
 & (0.0347) & (0.0097) & (0.0031) & (0.0154) & (0.006)\\ \hline
\end{tabular}
\caption{Mean and standard deviation (in parentheses) of the performance of each source task transferred to the target task Emory Emotion Recognition for each percentage of the target task data on BERT.}
\end{table*}

\begin{table*}
\centering
\begin{tabular}{|l|c|c|c|c|c|c|}
\hline
Percentage of Target Task Data & 20\% & 40\% & 60\% & 80\% & 100\%\\ \hline
\multicolumn{6}{|c|}{Target Task: Reading Comprehension on BERT}\\ \hline
No Task Transfer Reading Comprehension & 0.4826 & 0.5917 & 0.6146 & 0.6115 & 0.6388\\
 & (0.0317) & (0.0362) & (0.0158) & (0.0355) & (0.0393)\\ \hline
Source Task: Personality Detection & 0.4456 & 0.5521 & 0.5743 & 0.601 & 0.6124\\
 & (0.013) & (0.0137) & (0.0257) & (0.0104) & (0.0409)\\ \hline
Source Task: Relation Extraction & 0.438 & 0.5089 & 0.5244 & 0.5776 & 0.5719\\
 & (0.0136) & (0.0154) & (0.0251) & (0.0185) & (0.0346)\\ \hline
Source Task: Emory Emotion Recognition & 0.4498 & 0.5339 & 0.5782 & 0.5957 & 0.6369\\
 & (0.0031) & (0.0374) & (0.0199) & (0.0254) & (0.0398)\\ \hline
Source Task: Question Answering & 0.4634 & 0.5042 & 0.5258 & 0.5591 & 0.5889\\
 & (0.0008) & (0.0272) & (0.0217) & (0.0228) & (0.0323)\\ \hline
Source Task: Character Identification & 0.4926 & 0.556 & 0.5775 & 0.6019 & 0.6341\\
 & (0.018) & (0.0267) & (0.0292) & (0.0246) & (0.0327)\\ \hline
Source Task: Meld Emotion Recognition & 0.4895 & 0.5569 & 0.5595 & 0.6064 & 0.6108\\
 & (0.0391) & (0.0363) & (0.0207) & (0.0229) & (0.0362)\\ \hline
Source Task: Masked Language Modeling & 0.4826 & 0.5917 & 0.6146 & 0.6115 & 0.6388\\
 & (0.0317) & (0.0362) & (0.0158) & (0.0355) & (0.0393)\\ \hline
\end{tabular}
\caption{Mean and standard deviation (in parentheses) of the performance of each source task transferred to the target task Reading Comprehension for each percentage of the target task data on BERT.}
\end{table*}

\begin{table*}
\centering
\begin{tabular}{|l|c|c|c|c|c|c|}
\hline
Percentage of Target Task Data & 20\% & 40\% & 60\% & 80\% & 100\%\\ \hline
\multicolumn{6}{|c|}{Target Task: Question Answering on BERT}\\ \hline
No Task Transfer Question Answering & 0.2236 & 0.2782 & 0.3189 & 0.3283 & 0.3527\\
 & (0.0139) & (0.0168) & (0.0072) & (0.0069) & (0.0091)\\ \hline
Source Task: Personality Detection & 0.2343 & 0.277 & 0.3013 & 0.3395 & 0.3496\\
 & (0.0138) & (0.008) & (0.0137) & (0.0096) & (0.0146)\\ \hline
Source Task: Relation Extraction & 0.218 & 0.2649 & 0.31 & 0.3239 & 0.3477\\
 & (0.0127) & (0.0151) & (0.0232) & (0.0123) & (0.0197)\\ \hline
Source Task: Emory Emotion Recognition & 0.2406 & 0.2637 & 0.3023 & 0.3287 & 0.3528\\
 & (0.0092) & (0.0114) & (0.0128) & (0.0105) & (0.0165)\\ \hline
Source Task: Reading Comprehension & 0.2122 & 0.2709 & 0.2926 & 0.3223 & 0.3397\\
 & (0.0093) & (0.0211) & (0.0093) & (0.0094) & (0.0121)\\ \hline
Source Task: Character Identification & 0.2054 & 0.2597 & 0.2957 & 0.3202 & 0.3284\\
 & (0.008) & (0.013) & (0.0172) & (0.0148) & (0.0121)\\ \hline
Source Task: Meld Emotion Recognition & 0.2231 & 0.2723 & 0.3072 & 0.3189 & 0.3554\\
 & (0.0144) & (0.0011) & (0.0093) & (0.0096) & (0.0105)\\ \hline
Source Task: Masked Language Modeling & 0.2236 & 0.2782 & 0.3189 & 0.3283 & 0.3527\\
 & (0.0139) & (0.0168) & (0.0072) & (0.0069) & (0.0091)\\ \hline
\end{tabular}
\caption{Mean and standard deviation (in parentheses) of the performance of each source task transferred to the target task Question Answering for each percentage of the target task data on BERT.}
\end{table*}

\begin{table*}
\centering
\begin{tabular}{|l|c|c|c|c|c|c|}
\hline
Percentage of Target Task Data & 20\% & 40\% & 60\% & 80\% & 100\%\\ \hline
\multicolumn{6}{|c|}{Target Task: Meld Emotion Recognition on BERT}\\ \hline
No Task Transfer Meld Emotion Recognition & 0.4808 & 0.4985 & 0.5204 & 0.5309 & 0.5408\\
 & (0.0195) & (0.0136) & (0.0217) & (0.0102) & (0.0007)\\ \hline
Source Task: Personality Detection & 0.4822 & 0.4985 & 0.5225 & 0.5319 & 0.5462\\
 & (0.0128) & (0.0105) & (0.007) & (0.0127) & (0.0064)\\ \hline
Source Task: Relation Extraction & 0.3882 & 0.443 & 0.4898 & 0.4944 & 0.5139\\
 & (0.0235) & (0.036) & (0.0026) & (0.0029) & (0.0105)\\ \hline
Source Task: Emory Emotion Recognition & 0.5256 & 0.5289 & 0.532 & 0.5463 & 0.5638\\
 & (0.0122) & (0.0083) & (0.011) & (0.0047) & (0.0106)\\ \hline
Source Task: Reading Comprehension & 0.4142 & 0.484 & 0.5157 & 0.5139 & 0.5436\\
 & (0.0531) & (0.0054) & (0.0063) & (0.0101) & (0.0072)\\ \hline
Source Task: Question Answering & 0.4196 & 0.4804 & 0.5076 & 0.511 & 0.5279\\
 & (0.0422) & (0.0174) & (0.0033) & (0.0064) & (0.0082)\\ \hline
Source Task: Character Identification & 0.3577 & 0.4304 & 0.4639 & 0.482 & 0.5073\\
 & (0.0066) & (0.0438) & (0.0455) & (0.0317) & (0.0173)\\ \hline
Source Task: Masked Language Modeling & 0.4808 & 0.4985 & 0.5204 & 0.5309 & 0.5408\\
 & (0.0195) & (0.0136) & (0.0217) & (0.0102) & (0.0007)\\ \hline
\end{tabular}
\caption{Mean and standard deviation (in parentheses) of the performance of each source task transferred to the target task Meld Emotion Recognition for each percentage of the target task data on BERT.}
\end{table*}

\begin{table*}
\centering
\begin{tabular}{|l|c|c|c|c|c|c|}
\hline
Percentage of Target Task Data & 20\% & 40\% & 60\% & 80\% & 100\%\\ \hline
\multicolumn{6}{|c|}{Target Task: Character Identification on BERT}\\ \hline
No Task Transfer Character Identification & 0.8586 & 0.8703 & 0.8726 & 0.8789 & 0.8786\\
 & (0.0034) & (0.0038) & (0.0022) & (0.0033) & (0.0018)\\ \hline
Source Task: Personality Detection & 0.8564 & 0.8774 & 0.8794 & 0.8787 & 0.8849\\
 & (0.0032) & (0.0024) & (0.0011) & (0.0028) & (0.0013)\\ \hline
Source Task: Relation Extraction & 0.8451 & 0.8622 & 0.8687 & 0.8762 & 0.8795\\
 & (0.0047) & (0.0052) & (0.0027) & (0.0029) & (0.0026)\\ \hline
Source Task: Emory Emotion Recognition & 0.8528 & 0.8703 & 0.8768 & 0.8815 & 0.8813\\
 & (0.0053) & (0.0025) & (0.0038) & (0.0042) & (0.0032)\\ \hline
Source Task: Reading Comprehension & 0.8551 & 0.8679 & 0.8755 & 0.8739 & 0.8846\\
 & (0.0085) & (0.0025) & (0.0032) & (0.0067) & (0.0027)\\ \hline
Source Task: Question Answering & 0.8498 & 0.8644 & 0.8712 & 0.8717 & 0.8786\\
 & (0.0033) & (0.0056) & (0.0029) & (0.005) & (0.001)\\ \hline
Source Task: Meld Emotion Recognition & 0.8595 & 0.8732 & 0.876 & 0.8847 & 0.8829\\
 & (0.0062) & (0.0045) & (0.0026) & (0.0021) & (0.0061)\\ \hline
Source Task: Masked Language Modeling & 0.8586 & 0.8703 & 0.8726 & 0.8789 & 0.8786\\
 & (0.0034) & (0.0038) & (0.0022) & (0.0033) & (0.0018)\\ \hline
\end{tabular}
\caption{Mean and standard deviation (in parentheses) of the performance of each source task transferred to the target task Character Identification for each percentage of the target task data on BERT.}
\end{table*}

\begin{table*}
\centering
\begin{tabular}{|l|c|c|c|c|c|c|}
\hline
Percentage of Target Task Data & 20\% & 40\% & 60\% & 80\% & 100\%\\ \hline
\multicolumn{6}{|c|}{Target Task: Personality Detection on T5}\\ \hline    
No Task Transfer Personality Detection & 0.5273 & 0.5588 & 0.5709 & 0.5733 & 0.5879\\
 & (0.0247) & (0.0179) & (0.0079) & (0.0052) & (0.0073)\\ \hline
Source Task: Relation Extraction & 0.5364 & 0.5588 & 0.5558 & 0.5503 & 0.5412\\
 & (0.02) & (0.0164) & (0.0089) & (0.014) & (0.0216)\\ \hline
Source Task: Emory Emotion Recognition & 0.4939 & 0.5618 & 0.577 & 0.5552 & 0.5655\\
 & (0.0194) & (0.0129) & (0.006) & (0.0193) & (0.0155)\\ \hline
Source Task: Reading Comprehension & 0.5164 & 0.5352 & 0.5491 & 0.5618 & 0.5606\\
 & (0.0239) & (0.0052) & (0.0107) & (0.0311) & (0.0284)\\ \hline
Source Task: Question Answering & 0.5133 & 0.5315 & 0.5606 & 0.5539 & 0.5485\\
 & (0.0415) & (0.0228) & (0.0166) & (0.0052) & (0.0111)\\ \hline
Source Task: Character Identification & 0.5194 & 0.5509 & 0.5461 & 0.5855 & 0.5594\\
 & (0.0446) & (0.0171) & (0.0073) & (0.0136) & (0.0009)\\ \hline
Source Task: Meld Emotion Recognition & 0.5273 & 0.5588 & 0.5709 & 0.5733 & 0.5879\\
 & (0.0247) & (0.0179) & (0.0079) & (0.0052) & (0.0073)\\ \hline
\end{tabular}
\caption{Mean and standard deviation (in parentheses) of the performance of each source task transferred to the target task Personality Detection for each percentage of the target task data on T5.}
\end{table*}

\begin{table*}
\centering
\begin{tabular}{|l|c|c|c|c|c|c|}
\hline
Percentage of Target Task Data & 20\% & 40\% & 60\% & 80\% & 100\%\\ \hline
\multicolumn{6}{|c|}{Target Task: Relation Extraction on T5}\\ \hline
No Task Transfer Relation Extraction & 0.448 & 0.4921 & 0.5078 & 0.5278 & 0.5278\\
 & (0.024) & (0.0107) & (0.0095) & (0.0183) & (0.0018)\\ \hline
Source Task: Personality Detection & 0.4236 & 0.5069 & 0.504 & 0.5355 & 0.5404\\
 & (0.0435) & (0.0088) & (0.0087) & (0.0221) & (0.0172)\\ \hline
Source Task: Emory Emotion Recognition & 0.4375 & 0.4994 & 0.5211 & 0.5419 & 0.5489\\
 & (0.0292) & (0.0031) & (0.0178) & (0.0139) & (0.0114)\\ \hline
Source Task: Reading Comprehension & 0.4338 & 0.4885 & 0.5224 & 0.5348 & 0.5462\\
 & (0.0313) & (0.0166) & (0.0127) & (0.0117) & (0.0118)\\ \hline
Source Task: Question Answering & 0.4157 & 0.4967 & 0.523 & 0.5361 & 0.5592\\
 & (0.0115) & (0.0071) & (0.009) & (0.0179) & (0.0092)\\ \hline
Source Task: Character Identification & 0.4524 & 0.487 & 0.5209 & 0.5455 & 0.5623\\
 & (0.012) & (0.0142) & (0.0147) & (0.0142) & (0.0089)\\ \hline
Source Task: Meld Emotion Recognition & 0.448 & 0.4921 & 0.5078 & 0.5278 & 0.5278\\
 & (0.024) & (0.0107) & (0.0095) & (0.0183) & (0.0018)\\ \hline
\end{tabular}
\caption{Mean and standard deviation (in parentheses) of the performance of each source task transferred to the target task Relation Extraction for each percentage of the target task data on T5.}
\end{table*}

\begin{table*}
\centering
\begin{tabular}{|l|c|c|c|c|c|c|}
\hline
Percentage of Target Task Data & 20\% & 40\% & 60\% & 80\% & 100\%\\ \hline
\multicolumn{6}{|c|}{Target Task: Emory Emotion Recognition on T5}\\ \hline
No Task Transfer Emory Emotion Recognition & 0.3465 & 0.3644 & 0.3582 & 0.3655 & 0.3729\\
 & (0.0125) & (0.0149) & (0.0065) & (0.0051) & (0.0126)\\ \hline
Source Task: Personality Detection & 0.3109 & 0.3434 & 0.3647 & 0.3652 & 0.3665\\
 & (0.0135) & (0.0097) & (0.0031) & (0.0097) & (0.0013)\\ \hline
Source Task: Relation Extraction & 0.3159 & 0.3294 & 0.3443 & 0.3594 & 0.3682\\
 & (0.0038) & (0.0086) & (0.0169) & (0.0059) & (0.0078)\\ \hline
Source Task: Reading Comprehension & 0.2983 & 0.3378 & 0.349 & 0.3632 & 0.3545\\
 & (0.0127) & (0.0017) & (0.0141) & (0.013) & (0.0023)\\ \hline
Source Task: Question Answering & 0.296 & 0.335 & 0.354 & 0.3657 & 0.3624\\
 & (0.0158) & (0.0056) & (0.0051) & (0.0013) & (0.0075)\\ \hline
Source Task: Character Identification & 0.3181 & 0.3378 & 0.3555 & 0.3532 & 0.3672\\
 & (0.0102) & (0.0167) & (0.0047) & (0.0156) & (0.0034)\\ \hline
Source Task: Meld Emotion Recognition & 0.3465 & 0.3644 & 0.3582 & 0.3655 & 0.3729\\
 & (0.0125) & (0.0149) & (0.0065) & (0.0051) & (0.0126)\\ \hline
\end{tabular}
\caption{Mean and standard deviation (in parentheses) of the performance of each source task transferred to the target task Emory Emotion Recognition for each percentage of the target task data on T5.}
\end{table*}

\begin{table*}
\centering
\begin{tabular}{|l|c|c|c|c|c|c|}
\hline
Percentage of Target Task Data & 20\% & 40\% & 60\% & 80\% & 100\%\\ \hline
\multicolumn{6}{|c|}{Target Task: Reading Comprehension on T5}\\ \hline
No Task Transfer Reading Comprehension & 0.4806 & 0.5579 & 0.5736 & 0.5864 & 0.6062\\
 & (0.0321) & (0.0194) & (0.035) & (0.0186) & (0.0079)\\ \hline
Source Task: Personality Detection & 0.4872 & 0.5584 & 0.5736 & 0.5959 & 0.6073\\
 & (0.0156) & (0.028) & (0.0333) & (0.0056) & (0.0097)\\ \hline
Source Task: Relation Extraction & 0.4828 & 0.5654 & 0.5773 & 0.5938 & 0.6079\\
 & (0.0336) & (0.0057) & (0.0067) & (0.0144) & (0.0069)\\ \hline
Source Task: Emory Emotion Recognition & 0.5185 & 0.5727 & 0.5943 & 0.6068 & 0.6146\\
 & (0.0105) & (0.0222) & (0.0047) & (0.0174) & (0.0119)\\ \hline
Source Task: Question Answering & 0.5565 & 0.5825 & 0.5987 & 0.6147 & 0.6242\\
 & (0.0006) & (0.0076) & (0.0014) & (0.0157) & (0.0072)\\ \hline
Source Task: Character Identification & 0.5665 & 0.5989 & 0.6105 & 0.6116 & 0.6211\\
 & (0.0134) & (0.0175) & (0.0035) & (0.0199) & (0.0057)\\ \hline
Source Task: Meld Emotion Recognition & 0.4806 & 0.5579 & 0.5736 & 0.5864 & 0.6062\\
 & (0.0321) & (0.0194) & (0.035) & (0.0186) & (0.0079)\\ \hline
\end{tabular}
\caption{Mean and standard deviation (in parentheses) of the performance of each source task transferred to the target task Reading Comprehension for each percentage of the target task data on T5.}
\end{table*}

\begin{table*}
\centering
\begin{tabular}{|l|c|c|c|c|c|c|}
\hline
Percentage of Target Task Data & 20\% & 40\% & 60\% & 80\% & 100\%\\ \hline
\multicolumn{6}{|c|}{Target Task: Question Answering on T5}\\ \hline
No Task Transfer Question Answering & 0.3467 & 0.3618 & 0.3689 & 0.3829 & 0.3879\\
 & (0.0208) & (0.0198) & (0.0125) & (0.0115) & (0.0139)\\ \hline
Source Task: Personality Detection & 0.3706 & 0.3801 & 0.3905 & 0.3931 & 0.4032\\
 & (0.0059) & (0.0091) & (0.0045) & (0.0029) & (0.003)\\ \hline
Source Task: Relation Extraction & 0.3375 & 0.3526 & 0.369 & 0.3745 & 0.3803\\
 & (0.0111) & (0.0107) & (0.0122) & (0.0107) & (0.0054)\\ \hline
Source Task: Emory Emotion Recognition & 0.3685 & 0.3753 & 0.3928 & 0.3985 & 0.3975\\
 & (0.0039) & (0.0051) & (0.0015) & (0.0004) & (0.0104)\\ \hline
Source Task: Reading Comprehension & 0.3851 & 0.3882 & 0.3946 & 0.4031 & 0.4099\\
 & (0.0063) & (0.0026) & (0.0016) & (0.0081) & (0.0045)\\ \hline
Source Task: Character Identification & 0.3639 & 0.3644 & 0.3831 & 0.3948 & 0.4008\\
 & (0.0068) & (0.0044) & (0.0067) & (0.0037) & (0.0076)\\ \hline
Source Task: Meld Emotion Recognition & 0.3467 & 0.3618 & 0.3689 & 0.3829 & 0.3879\\
 & (0.0208) & (0.0198) & (0.0125) & (0.0115) & (0.0139)\\ \hline
\end{tabular}
\caption{Mean and standard deviation (in parentheses) of the performance of each source task transferred to the target task Question Answering for each percentage of the target task data on T5.}
\end{table*}

\begin{table*}
\centering
\begin{tabular}{|l|c|c|c|c|c|c|}
\hline
Percentage of Target Task Data & 20\% & 40\% & 60\% & 80\% & 100\%\\ \hline
\multicolumn{6}{|c|}{Target Task: Meld Emotion Recognition on T5}\\ \hline
No Task Transfer Meld Emotion Recognition & 0.4997 & 0.5121 & 0.5206 & 0.5252 & 0.5238\\
 & (0.0103) & (0.0095) & (0.0083) & (0.0161) & (0.0111)\\ \hline
Source Task: Personality Detection & 0.4902 & 0.485 & 0.5198 & 0.5193 & 0.528\\
 & (0.0118) & (0.0029) & (0.0065) & (0.0061) & (0.0177)\\ \hline
Source Task: Relation Extraction & 0.4891 & 0.5033 & 0.5166 & 0.5188 & 0.5227\\
 & (0.0123) & (0.0189) & (0.0111) & (0.0053) & (0.0125)\\ \hline
Source Task: Emory Emotion Recognition & 0.4896 & 0.5049 & 0.5221 & 0.5282 & 0.525\\
 & (0.0091) & (0.0139) & (0.0025) & (0.0046) & (0.0023)\\ \hline
Source Task: Reading Comprehension & 0.4759 & 0.4987 & 0.5245 & 0.5321 & 0.5316\\
 & (0.0037) & (0.0072) & (0.0027) & (0.012) & (0.0058)\\ \hline
Source Task: Question Answering & 0.4791 & 0.5061 & 0.5125 & 0.5211 & 0.5329\\
 & (0.014) & (0.0162) & (0.0074) & (0.0058) & (0.0042)\\ \hline
Source Task: Character Identification & 0.5003 & 0.5101 & 0.5188 & 0.527 & 0.5229\\
 & (0.0151) & (0.0072) & (0.0062) & (0.0031) & (0.0064)\\ \hline
\end{tabular}
\caption{Mean and standard deviation (in parentheses) of the performance of each source task transferred to the target task Meld Emotion Recognition for each percentage of the target task data on T5.}
\end{table*}

\begin{table*}
\centering
\begin{tabular}{|l|c|c|c|c|c|c|}
\hline
Percentage of Target Task Data & 20\% & 40\% & 60\% & 80\% & 100\%\\ \hline
\multicolumn{6}{|c|}{Target Task: Character Identification on T5}\\ \hline
No Task Transfer Character Identification & 0.5187 & 0.6144 & 0.6414 & 0.6786 & 0.6418\\
 & (0.0359) & (0.0092) & (0.011) & (0.0081) & (0.0153)\\ \hline
Source Task: Personality Detection & 0.4901 & 0.6019 & 0.6665 & 0.6929 & 0.6864\\
 & (0.0534) & (0.019) & (0.0251) & (0.0142) & (0.0318)\\ \hline
Source Task: Relation Extraction & 0.4967 & 0.6058 & 0.6369 & 0.68 & 0.6795\\
 & (0.0523) & (0.032) & (0.0163) & (0.0038) & (0.0022)\\ \hline
Source Task: Emory Emotion Recognition & 0.5737 & 0.6228 & 0.6553 & 0.6535 & 0.7033\\
 & (0.0121) & (0.0029) & (0.0305) & (0.0045) & (0.0103)\\ \hline
Source Task: Reading Comprehension & 0.5988 & 0.6246 & 0.6883 & 0.6998 & 0.7108\\
 & (0.0062) & (0.001) & (0.0095) & (0.0071) & (0.0058)\\ \hline
Source Task: Question Answering & 0.5974 & 0.643 & 0.6542 & 0.675 & 0.6974\\
 & (0.0044) & (0.0241) & (0.0062) & (0.0146) & (0.025)\\ \hline
Source Task: Meld Emotion Recognition & 0.5187 & 0.6144 & 0.6414 & 0.6786 & 0.6418\\
 & (0.0359) & (0.0092) & (0.011) & (0.0081) & (0.0153)\\ \hline
\end{tabular}
\caption{Mean and standard deviation (in parentheses) of the performance of each source task transferred to the target task Character Identification for each percentage of the target task data on T5.}
\end{table*}

\end{document}